\documentclass[runningheads]{llncs}

\usepackage{eccv}

\usepackage{eccvabbrv}

\usepackage{multirow}
\usepackage{graphicx}
\usepackage{booktabs}
\usepackage{color}
\usepackage{amssymb}
\usepackage{xcolor}
\usepackage{colortbl}
\usepackage{wrapfig}

\usepackage[accsupp]{axessibility}  

\definecolor{mygray}{gray}{0.9}

\usepackage[pagebackref,breaklinks,colorlinks,citecolor=eccvblue]{hyperref}

\usepackage{orcidlink}

\begin{document}

\title{An Economic Framework for 6-DoF \\ Grasp Detection}


\author{Xiao-Ming Wu\inst{1, 3, \dagger, \&\orcidlink{0000-0003-1115-8551}} \and
Jia-Feng Cai\inst{1, 3, \&\orcidlink{0009-0004-4203-2745}} \and
Jian-Jian Jiang\inst{1, 3\orcidlink{0009-0009-2324-639X}} \and
Dian Zheng\inst{1, 3\orcidlink{0009-0009-1038-0223}} \and
Yi-Lin Wei\inst{1, 3\orcidlink{0009-0004-9210-7370}} \and
Wei-Shi Zheng\inst{1, 2, 3, *\orcidlink{0000-0001-8327-0003}}}

\authorrunning{X.M. Wu et al.}

\institute{School of Computer Science and Engineering, Sun Yat-sen University, China \and
Peng Cheng Laboratory, Shenzhen, China \and
Key Laboratory of Machine Intelligence and Advanced Computing, Ministry of Education, China \\
\email{\{wuxm65, caijf23, jiangjj35, zhengd35, weiylin5\}@mail2.sysu.edu.cn, wszheng@ieee.org}}

\begingroup
\renewcommand{\thefootnote}{\relax}
\footnotetext{$\dagger:$ project lead, \&: equal key contributions, *: corresponding author.}
\endgroup

\maketitle
\begin{abstract}
  Robotic grasping in clutters is a fundamental task in robotic manipulation. In this work, we propose an economic framework for 6-DoF grasp detection, aiming to economize the resource cost in training and meanwhile maintain effective grasp performance. To begin with, we discover that the dense supervision is the bottleneck of current SOTA methods that severely encumbers the entire training overload, meanwhile making the training difficult to converge. To solve the above problem, we first propose an economic supervision paradigm for efficient and effective grasping. This paradigm includes a well-designed supervision selection strategy, selecting key labels basically without ambiguity, and an economic pipeline to enable the training after selection. Furthermore, benefit from the economic supervision, we can focus on a specific grasp, and thus we devise a focal representation module, which comprises an interactive grasp head and a composite score estimation to generate the specific grasp more accurately. Combining all together, the \textbf{EconomicGrasp} framework is proposed. Our extensive experiments show that EconomicGrasp surpasses the SOTA grasp method by about \textbf{3AP} on average, and with extremely low resource cost, for about \textbf{1/4} training time cost, \textbf{1/8} memory cost and \textbf{1/30} storage cost. Our code is available at \url{https://github.com/iSEE-Laboratory/EconomicGrasp}.
\end{abstract}

\vspace{-3mm}
\section{Introduction}
\label{sec:intro}
\vspace{-2mm}

\begin{figure}[t]
  \centering
  \includegraphics[width=1\textwidth]{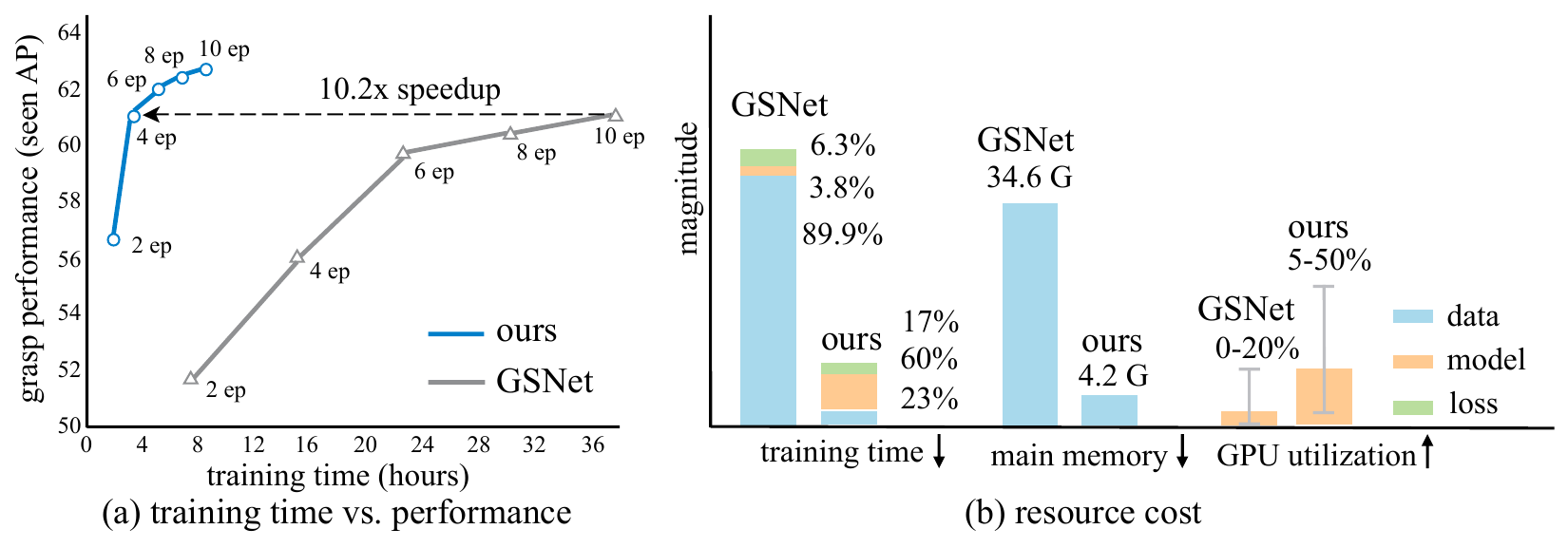}
  \caption{\textbf{Economic supervision vs. dense supervision.} In our economic framework, the resource cost is minimal and the training is easy to converge. Moreover, with our well design for economic supervision, our framework achieves better performance than the SOTA dense supervision method GSNet\cite{wang2021graspness}. "ep" means epochs. All the costs are tested in an empty machine with one NVIDIA RTX3090 GPU for fair. The results are trained and tested with GraspNet-1Billion\cite{fang2020graspnet} on Kinect data.}
  \label{fig:motivation}
\vspace{-6mm}
\end{figure}

\noindent As a fundamental task in robotic manipulations, 6-DoF grasp detection \cite{ten2017grasp, fang2020graspnet, breyer2021volumetric, fang2023anygrasp} achieves rapid development in recent years, which has broad applications in many real-world scenarios, such as picking\cite{correll2016analysis}, assembling\cite{suarez2018can}, fruit harvesting\cite{zhou2022intelligent} and house serving like cooking and cleaning\cite{fu2024mobile}.

Recently, with the help of the real-world grasp dataset GraspNet-1Billion\cite{fang2020graspnet}, which has abundant labels as supervision (we call it \textbf{dense supervision}, which is in billion scale), 6-DoF grasp detection methods\cite{fang2020graspnet, gou2021rgb, wang2021graspness, ma2023towards, fang2023anygrasp} achieve great grasping performance. However, this type of supervision also occurs some problems. It 1) has huge resource costs in model training, 2) makes the training difficult to learn and converge. As shown in Fig. \ref{fig:motivation} (a), this supervision is somehow too abundant for the model to train, which causes slow and hard convergence. Furthermore, as shown in Fig. \ref{fig:motivation} (b), the data processing time of the state-of-the-art method GSNet\cite{wang2021graspness}, which is under dense supervision, is very long (about 9x of the sum of model training time and loss calculation time). Moreover, nearly 100 million labels will be loaded in each batch to train the model, resulting in high memory cost (34.6G) and low GPU utilization (20\%). Prior to GraspNet-1Billion\cite{fang2020graspnet}, due to supervision-limited, many grasping methods are designed to learn in \textbf{sparse supervision}\cite{ten2017grasp, liang2019pointnetgpd, ni2020pointnet++, qin2020s4g, breyer2021volumetric} (in million scale). Although they may have fewer training resources or fast converge speed, they have limited grasp performance due to the inadequate training data. Therefore, how to enable economic grasping with low resource costs and effective grasp performance is a meaningful but rarely explored area in 6-DoF grasp detection, which can promote the extension of the research area and support wide applications in resource-constrained environments.

In this work, we propose an economic framework for 6-DoF grasp detection, which aims to \textbf{economize the resource costs in training and meanwhile maintain effective grasp performance}. Firstly, we analyze the reason which causes the gap between sparse and dense supervision methods. We follow the previous sparse supervision methods, gradually "modernizing" it, to see whether it is the module or loss designs that lead the gap. However, the answer is NO and there still exist other problems in sparse supervision. Secondly, we conduct variance analysis in labels, and find that the ambiguity problem is the "chief culprit" causing this gap. Therefore, we propose an economic supervision paradigm for efficient and effective grasping, including a well-designed supervision selection strategy and a economic pipeline to enable training after selection. Thirdly, economic supervision affords us the opportunity to focus on the learning of a specific grasp. Toward this end, we design a focal representation module to enable it, which includes an interactive grasp head to learn discriminate features for a specific grasp, and a composite score for more accurate score estimation. Combining all together, the \textbf{EconomicGrasp} is proposed, which is resource-friendly and performance-effective for robotic grasping.

Extensive experiments show that EconomicGrasp has great grasp performance, surpassing the state-of-the-art grasp method with 3AP on average. More importantly, our framework is economic, which has extremely low resource costs comparing to state-of-the-art grasp method, for about 1/4 training time cost, 1/8 memory cost and 1/30 storage cost. Our code is available at \url{https://github.com/iSEE-Laboratory/EconomicGrasp}.

\vspace{-3mm}
\section{Revisiting 6-DoF Grasp Detection}
\vspace{-2mm}
\label{sec:related_works}
\subsubsection{Task Definition.} 
\begin{figure}[t]
  \centering
  \includegraphics[width=0.9\textwidth]{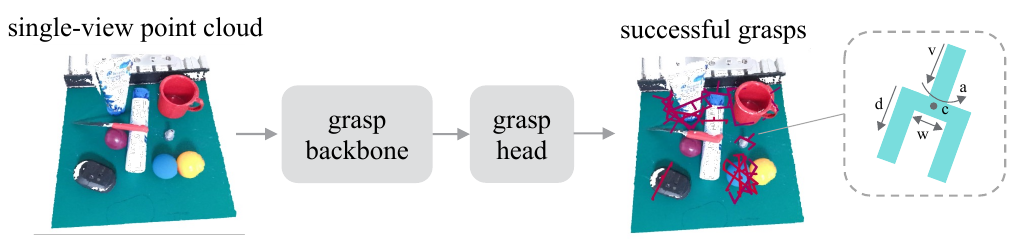}
  \vspace{-2mm}
  \caption{\textbf{Task definition} and the grasp pose. The input for this task is the single-view point cloud from the depth camera and the model aims to output the successful 6-DoF grasp poses for the input scene.}
  \label{fig:task}
\vspace{-6mm}
\end{figure}

Robotic grasping is a fundamental skill for robotic manipulation\cite{zhou2022intelligent, fu2024mobile, zhou2024mitigating}, which can be divided into two basic processes, perception and planing\cite{quigley2009ros}. The robot first detects where to grasp (grasp detection), then executes the control algorithm to move to the target position. 6-DoF grasp detection, different from the methods of detecting contact points\cite{saxena2008robotic, le2010learning} or rectangles\cite{jiang2011efficient, lenz2015deep, redmon2015real, morrison2018closing, yu2022egnet}, is to generate 6-DoF grasp poses based on the visual input\cite{ten2017grasp, liang2019pointnetgpd, mousavian20196, eppner2021acronym, sundermeyer2021contact, zhao2021regnet}, which consists of the 3D positions with its 3D rotations. Formally specking, giving the point cloud $\mathbf{C} \in \mathbb{R}^{n\times 3}$ as input ($n$ is the number of input points), our goal is to learn how to generate 6-DoF grasps that can successfully grab the object. Following the setting of previous works\cite{fang2020graspnet, wang2021graspness, chen2023grasp, liu2024simulating, fang2023anygrasp}, we represent the 6-DoF grasp as $ \mathbf{G}=[\mathbf{c}, v, a, d, w, s]$,
where $\mathbf{c} \in \mathbb{R}^{3}$ is the central point of the grasp, $v$ is an integer ranging from $1$ to $300$, standing for different approaching directions of the grasp view, $a$ is an integer ranging from $1$ to $12$, representing different angles of the 2D in-plane rotation, $d$ is also an integer ranging from $1$ to $4$, indicating the grasp depth, $w \in \mathbb{R}$ is the grasp width and $s \in \mathbb{R}$ is the grasp score describing the grasp quality. For better comprehension, we demonstrate the task definition and the 6-DoF grasp pose in Fig. \ref{fig:task}. More details can be found in \cite{fang2020graspnet}. To be noted that there are also other types of grippers like dexterous hand\cite{liu2020deep,wang2023dexgraspnet, huang2023diffusion, xu2023unidexgrasp,xu2024dexterous, wei2024grasp}, human hand\cite{jiang2021hand, wang2024single} and so on, here we mainly focus on the basic two-finger parallel gripper.

\vspace{-3mm}
\subsubsection{Development of 6-DoF Grasp Detection.} 
6-DoF grasp detection can trace back to many pioneering works\cite{fischinger2012empty, herzog2012template, kappler2015leveraging, ten2018using, gualtieri2016high, ten2017grasp} that aims to relax the restrict of the rectangle representation from three domains of freedom into the more relaxed ones. Among them, GPD\cite{ten2017grasp} is one of the most popular method that enables high success rate to grasp novel objects, which greatly promotes the research in this field.
Inspired by GPD\cite{ten2017grasp}, early 6-DoF grasp detection methods apply the sample-then-evaluation pipeline\cite{ten2017grasp, liang2019pointnetgpd, mousavian20196} to generate grasp. This pipeline first samples large amounts of grasp candidates from the input scene, then evaluates this candidate to see whether it is a successful grasp. However, this pipeline is highly time-consuming in inference, since that we should sample a large amount of grasp candidates and then put them into the evaluator to evaluate. This means that the evaluator (usually a neural network) should run several times to generate successful grasps. To enable efficient grasping, end-to-end methods are proposed \cite{sundermeyer2021contact, wei2021gpr, zhao2021regnet, cai2022real, shi2022symmetrygrasp, zheng2022vgpn, dai2023graspnerf, jauhri2023learning, zhai2023monograspnet, weng2023neural, liu2023rgbgrasp}, which directly generates successful 6-DoF grasps based on the visual input. 

Recent years, the large-scale real-world dataset GraspNet-1Billion \cite{asif2018graspnet} is proposed, which has dense grasp labels (in billion scale) and enables robust grasping in real world \cite{fang2020graspnet, gou2021rgb, wang2021graspness, ma2023towards, fang2023anygrasp, chen2023grasp, liu2024simulating}. However, the supervision in this pipeline may be too abundant, which may occur 1) huge resource costs in model training, 2) difficulty for the model to train and converge, as shown in Fig. \ref{fig:motivation}. Prior to the GraspNet-1Billion \cite{asif2018graspnet}, due to the data limit at that time, many methods are designed under sparse supervision\cite{ten2017grasp, liang2019pointnetgpd, ni2020pointnet++, qin2020s4g, breyer2021volumetric}. Although they may be resource-friendly or easy to converge, these methods obtain limited grasp performance. Meanwhile, few methods attempt to use partial labels in the large scale dataset GraspNet-1Billion \cite{asif2018graspnet} as supervision\cite{liu2022transgrasp, wang2023granet, li2021simultaneous, hoang2022context, qin2023rgb}. But they have neither explored how to conduct high-quality supervision selection for effective performance, nor attempted to pursue low resource costs when using partial supervision. Thus, they have limited grasp performance and not low resource costs. Thus, how to enable economic grasping (with low resource costs and effective grasp performance) is a meaningful but rarely explored area in 6-DoF grasp detection. Toward this end, in this work, we propose an economic framework, EcomonicGrasp, for 6-DoF grasp detection, aiming to economize the resource costs in training and meanwhile maintain effective grasp performance. 

\vspace{-3mm}
\section{The Economic Grasp Framework}
\vspace{-2mm}
\label{sec:economic_grasp}
Our economic grasp framework aims to enable effective grasping with low resource costs. First, we explore whether it is the module and loss designs causing the performance gap between the dense and sparse supervision methods. We find that they are part of the reason, but not the key factor (Sec. \ref{sec:vanilla}). Second, we conduct variance analysis in labels and discover that the ambiguity problem is the "chief culprit" causing this gap. Thus, we design an economic supervision paradigm to mitigate the ambiguity, comprising a supervision selection strategy and the corresponding training pipeline (Sec. \ref{sec:supervision}). Third, under economic supervision, we can focus on the learning of a specific grasp, which inspires us to further design the focal representation module (Sec. \ref{sec:representation}).

\vspace{-3mm}
\subsection{A Vanilla Grasp Framework}
\label{sec:vanilla}

\begin{wraptable}{r}{0.6\textwidth}
\vspace{-10mm}
\centering
\caption{We gradually "modernize" a simple method S4G\cite{qin2020s4g} (line 1) with the cutting-edge approaches, resulting in a vanilla grasp framework (line 4) for sparse supervision. However, its performance is still limited comparing to dense supervision methods. The performance are the seen results on GraspNet-1Billion dataset\cite{fang2020graspnet} in Kinect data.}
\resizebox{0.6\textwidth}{!}{
\begin{tabular}{c@{\hspace{1em}}c@{\hspace{1em}}c@{\hspace{1em}}cccc}
\toprule
PointNet++ & 3DConv & region & statistic & seen & similar & novel \\ 
\midrule
\checkmark & & & & 18.72 & 11.82 & 5.38 \\
\checkmark &  & \checkmark & & 37.38 & 29.50 & 10.23 \\
\checkmark &  & \checkmark & \checkmark & 39.63 & 31.72 & 11.09 \\
& \checkmark & \checkmark & \checkmark & 43.59 & 34.09 & 13.36 \\
\bottomrule
\end{tabular}}
\label{tab:strong_baseline}
\vspace{-4mm}
\end{wraptable}

Previous to the GraspNet-1Billion dataset\cite{fang2020graspnet}, there exists many methods training on sparse supervision. Thus, we firstly want to see whether it is the module and loss designs that cause the performance gap between the sparse supervision methods and dense supervision methods. To do so, we reduce the labels in GraspNet-1Billion dataset\cite{fang2020graspnet}, only keeping the best grasp for each point in each scene, 
And we implement a simple framework, which follows the PointNet++Grasp\cite{ni2020pointnet++} or S4G\cite{qin2020s4g} pipeline, using PointNet++\cite{qi2017pointnet++} as the backbone and directly predicting the grasp pose for each point, to enable the training under this supervision. However, as shown at the first line in Table \ref{tab:strong_baseline}, this simple baseline only has very limited performance. We further fuse some cutting-edge approaches into the simple baseline, gradually "modernizing" it into a better framework. First, region-based approaches are widely used in recent methods\cite{asif2018graspnet, zhao2021regnet, chen2023efficient, chen2023grasp}, so we apply the region-based method, cylinder grouping, used in\cite{asif2018graspnet, wang2021graspness}, into the model. Specifically, for each of the point feature outputted by the backbone, we first predict the grasp view, then group the points near the grasp center and along with the grasp view, forming a more discriminate feature to predict the remaining parts of the grasp pose. As shown in the second line in Table \ref{tab:strong_baseline}, region-based method largely improve the model performance. Furthermore, we apply the global statistic proposed in \cite{wang2021graspness} (the graspness identifier) as an extra task, to identify which points are suitable to grasp, which also improves the model performance. Finally, following many previous works\cite{breyer2021volumetric, wang2021graspness, chen2023grasp}, we use the 3D convolution\cite{graham20183d} instead of PointNet++\cite{qi2017pointnet++}, which also enhances the grasp performance. Combining all together, a vanilla grasp framework is proposed, which use sparse supervision to get satisfactory performance. However, there still a huge gap between the vanilla grasp framework and the SOTA method\cite{wang2021graspness} (achieve 61.19, 47.39, 19.01 performance in seen, similar and novel), where the only main difference between them are the supervision modes. To sum up, we have the following conclusion:
\begin{quote}
\textit{Cutting-edge approaches help to bridge the gap between sparse and dense supervision methods. But, a significant performance gap still remains.}
\end{quote}

\vspace{-3mm}
\subsection{Economic Supervision}
\label{sec:supervision}

\subsubsection{Ambiguity Problem.}
\begin{figure}[t]
  \centering
  \includegraphics[width=0.8\textwidth]{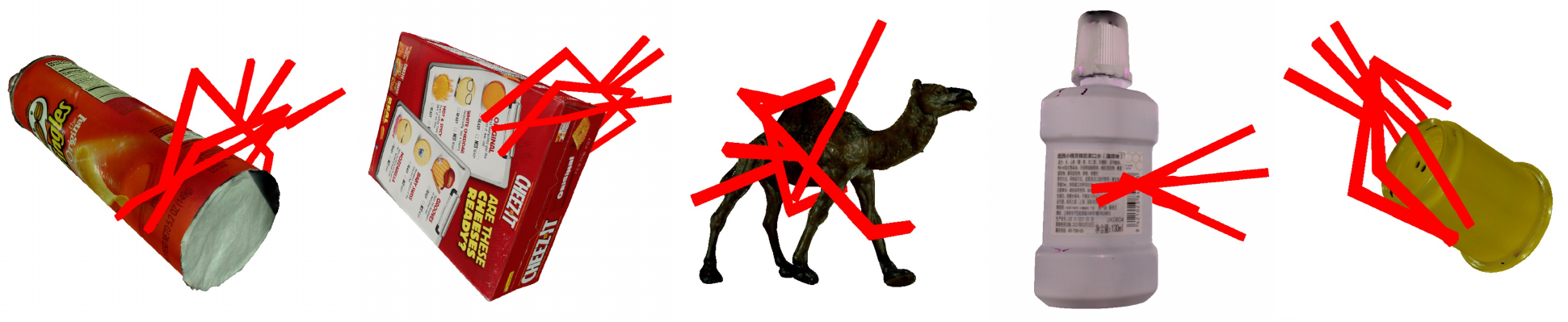}
  \vspace{-2mm}
  \caption{In specific points, there exist many good grasps that with \textbf{different poses}. If we reduce the supervision, it may cause ambiguity in the learning process.}
  \label{fig:ambiguity}
\vspace{-6mm}
\end{figure}

Now we further explore what problems cause the remain performance gap. An interesting idea comes to our mind when we visualize the good grasps selected in each objects. As shown in Fig. \ref{fig:ambiguity}, we observe that the good grasps in the specific points in different objects are in different poses (views, angles and depths are all different). In GraspNet-1Billion dataset\cite{fang2020graspnet}, it labels all the possible views, angles and depths ($300\times12\times4$) for each point in the object. If we attempt to reduce the labels in each point, it may only keep some of the probabilities within the point. In this case, similar points, especially in neighbor or the same points, may have different labels as supervision, which will confuse the neural networks training. We call this the \textbf{ambiguity problem}.


\begin{wraptable}{r}{0.4\textwidth}
\vspace{-10mm}
\centering
\caption{The standard deviation of good grasps within a point.}
\resizebox{0.4\textwidth}{!}{
\begin{tabular}{lccc}
\toprule
& view & angle & depth \\
\midrule
original & 46.81 & 3.18 & 0.81 \\
determine the view & -     & 0.22 & 0.05 \\
determine the angle & 13.45 & -    & 0.27 \\
determine the depth & 33.81 & 2.28 & -    \\
\bottomrule
\end{tabular}}
\label{tab:variance_good_grasps}
\vspace{-6mm}
\end{wraptable}

To further quantitatively verify this, we analyze the standard deviation of the view, angle and depth of the good grasps in each point. As shown in the first line in Table \ref{tab:variance_good_grasps}, the good grasp poses are actually diversified and may cause ambiguity problem when we reduce them. In addition, we can also see that grasp views are the most ambiguous part, which has the largest standard variance due to its wide range. Furthermore, we are curious that if we already determine the grasp view, angle or depth beforehand, how ambiguous will be of the remaining parts of the grasp pose. To this end, we test the standard deviation of the good grasps after determining the view, angle or depths in each point. The results are demonstrated from the two to four lines in Table \ref{tab:variance_good_grasps}. We can see that if the grasp view is determined beforehand, the angle and depth are also nearly fixed, indicating that the ambiguity problem is basically solved. In contrast, although determining the grasp angle or grasp depth can somehow makes the view less diverse, it is still variant. In summary, we can give the following conclusion:
\begin{quote}
\textit{The ambiguity problem exists when reducing the supervision, which causes the gap between sparse and dense supervision methods. Moreover, grasp view is the most important prat to mitigate this problem.}
\end{quote}

\vspace{-3mm}
\subsubsection{Economic Supervision Paradigm.}
Based on the above analysis, we recommend to keep all the view as supervision rather than only keeping one grasp in each point as in the vanilla grasp framework, which can largely mitigate the ambiguity problem. Under this consensus, we propose the economic supervision paradigm to implement this proposal, including the a well-designed supervision selection strategy to reduce supervision, and a pipeline to enable its training under this type of supervision.

First, we detail the procedure of the selection for economic supervision. The first step is grasp pose pruning. We keep the grasp view following the above recommend, and prune the grasp angle and grasp depth for each point. In detail, we only keep the view graspness\cite{wang2021graspness} and the best grasp for each view (300 grasps for each point instead of previous $300\times12\times4$ grasp poses for each point). Second, we collect the grasp labels of all the objects belonging to each scene to form the scene-level labels, instead of constructing them during training. This process will increase the storage size of the label set. But due to the grasp pose pruning at the first step, this increase is acceptable and will improve data loading speed in training. The final step is point pruning. Since we have already constructed the scene-level labels in the second step, we can further prune the labels in the scene points without graspable grasp labels (with friction coefficient\cite{nguyen1988constructing, ten2017grasp, liang2019pointnetgpd, fang2020graspnet} lower than 0.8 or with collision in the scene). This step will reduce the label set to 1/2 of the original. After the above three steps, we can reduce the label set from \textbf{55 GB to 1.6 GB}, becoming lower than 1/30 of the original, and meanwhile construct the scene-level labels in advance. Therefore, we can rapidly load the labels when training, which also has less memory and GPU costs, largely economizing the training overload, and meanwhile enabling fast and stable convergence. 

To enable the model to train under the above economic supervision, we just need to slightly revise the vanilla grasp framework describe in Sec. \ref{sec:vanilla}. First, we predict the view scores for each point, and then select the best view based on the predicted scores. Next, we generate the remaining parts of the grasp pose for the selected view. The grasp view training is based on the view graspness\cite{wang2021graspness}. Since we already keep a grasp for each view, we can use the corresponding label to supervise the learning of the remaining parts (angle, depth, width and score). Besides, since we also prune the labels in points without good grasps, the labels are hard to match all the input points (we may not have suitable labels close enough to supervise the training of the grasps of some the input points). Thus, we use a selective match loss, masking the loss for the grasps of input points without suitable labels and preventing back-propagating its loss.

\vspace{-3mm}
\subsection{Focal Representation under Economic Supervision}
After economic supervision, we have the chance to give comprehensive consideration to learn a specific grasp. Thus, we design the focal representation module, comprising an interactive grasp head and a composite score estimation module to learn the specific grasp more accurately.

\label{sec:representation}

\vspace{-3mm}
\subsubsection{Interactive Grasp Head.}
The insight of the interactive grasp head is that since we only keep one grasp for each view, we aim to facilitate more interaction among the feature, to be more aligned with the  specific grasp. Thus, we design a global and local interactive attentions\cite{vaswani2017attention} to achieve this goal.

First, the region-based method described in Sec. \ref{sec:vanilla} will output the feature that can represent the region near the point along a specific view, which basically contains the range of all possible grasps to this point along the selected view. However, since we only aim to learn a specific good grasp in this range under economic supervision, the region is still too broad for our task. Therefore, we design a global interactive attention to learn interaction within the region, which conducts the attention within the regions to construct a more discriminate feature for the learning the specific grasp. After that, following the previous object detection methods\cite{girshick2015fast, ren2015faster, wu2020rethinking}, we use multiple heads to learn the remaining parts of the grasp $r,d,w,s$ from different features, instead of directly learning them from a common feature space.

\begin{wrapfigure}{r}{0.35\textwidth}
\vspace{-8mm}
  \centering
  \includegraphics[width=0.32\textwidth]{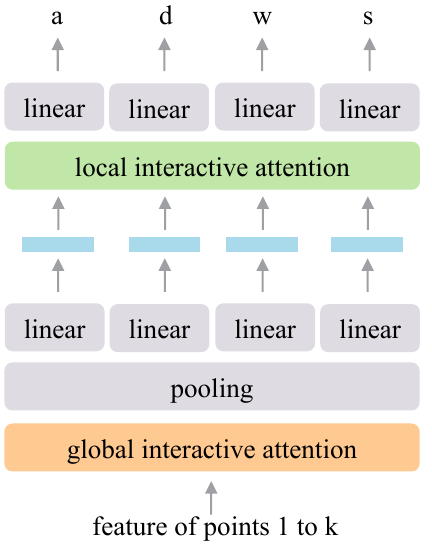}
  \caption{Interactive grasp head.}
  \label{fig:interactive_head}
\vspace{-10mm}
\end{wrapfigure}

Second, the remaining parts of the grasp are dependent, and this dependent relation is not determined beforehand. To be specific, sometimes when we have determined the angle, the depth naturally becomes fixed. However, this is not always the case. In some instances, establishing the depth first is better, which can result in a nearly-fixed angle. To learn this dependent relation, we use a local attention module to enable it, which executes the attention within the four grasp features. For clarity, we demonstrate the interactive grasp head in Fig. \ref{fig:interactive_head}. To be noted that the global interactive attention is conducted among the region points, following a pooling layer to fuse them into an unified feature, and the local interactive attention is executed among different parts of the grasp pose.

This design is mainly for economic supervision. In previous dense supervision methods, they directly predict the combination of angles and depths, which is hard to learn features focusing on all the grasp poses economically.

\vspace{-3mm}
\subsubsection{Composite Score Estimation.}
What's more, we want to predict the score more accurately for the specific grasp. Inspired by the idea from action quality assessment\cite{tang2020uncertainty}, which aims to predict a score to reflect the quality of the actions, we can model the score by classification. To be specific,  we model the score by simply $1.1-\mu$, where $\mu$ is the friction coefficient\cite{nguyen1988constructing, ten2017grasp, liang2019pointnetgpd, fang2020graspnet}. In this way, the scores will have six value (from 0 to 1 with $\bigtriangleup 0.2$) and can be learnt by a six-class classifier. For inference, naive approach is to predict the score using the value corresponding to the highest probability. However, this cannot fully leverage the advantage of learning the score distribution using classification. Instead, we use the composite score estimation, linearly combining the scores in each segment by the probability, which can be formulated as $
s=[0, 0.2, 0.4, 0.6, 0.8, 1]\cdot \mathbf{s} _{c}^{T}$, 
where $\mathbf{s}_{c}$ is the output of the score classifier. The composite score estimation is highly necessary for predicting accurate score, which is evaluated in Sec. \ref{sec:ablation}.

To be noted that this idea is also tailored for economic supervision. If we introduce this idea into the dense framework, the composite score estimation should transform the regression head into a classification head, this will make the angle-depth combination in dense supervision methods six times larger, which is highly uneconomical.

\vspace{-3mm}
\subsection{Framework Overall}
\begin{figure}[t]
  \centering
  \includegraphics[width=0.9\textwidth]{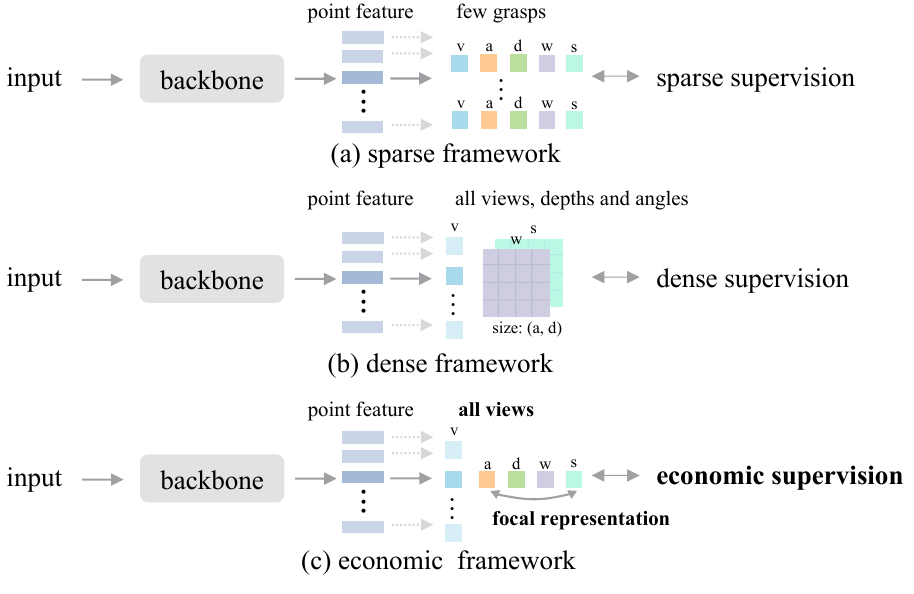}
  \vspace{-4mm}
  \caption{\textbf{Frameworks overview}. Benefit from the economic supervision and focal representation, our economic framework achieves effective performance with low costs. We highlight the main contributions of our paper in bold face.}
  \label{fig:framework}
\vspace{-4mm}
\end{figure}

The overall economic framework is described in Fig. \ref{fig:framework}. Previous sparse framework predicts one or few grasps for each point. However, this learning is somehow ambiguous since that the output just contains partial probabilities of all the good grasp, which results in limited performance, as analyzed in Sec.\ref{sec:supervision}. The dense framework will output and learn the combination of grasp angles and grasp depths, which has high resource costs and slow convergence speed, as shown in Fig. \ref{fig:motivation}. Our economic framework innovatively keeps
all views to mitigate the ambiguity problem, and design a focal representation module to generate the specific grasp more accurately.

\subsection{Dataset and Details}
\label{sec:exp_setting}
\subsubsection{Dataset.}
We use GraspNet-1Billion\cite{fang2020graspnet} for training and evaluation. GraspNet-1Billion\cite{fang2020graspnet} is a large-scale real-world dataset with $88$ objects and $190$ scenes captured by two cameras (Kniect and RealScenes), which can reflect the real-world grasp performance accurately\cite{wang2021graspness, fang2023anygrasp}. There are 100 scenes as the training set and 90 scenes to be the testing set. In addition, the testing set are further equally divided into three splits for seen, similar and novel objects respectively. To be noted that this is the only dense real-world dataset to fully analyze the performance and costs of our framework with other sparse or dense frameworks.

\vspace{-3mm}
\subsubsection{Metric.}
As for the metric, we adopt $\mathrm{AP}$ as the metric following GraspNet-1Billion\cite{fang2020graspnet}. In detail, we first generate 6-DoF grasps by our model, then evaluate the grasps with high scores by reconstructing the test scenes and calculating the force-closure metric\cite{nguyen1988constructing, ten2017grasp, liang2019pointnetgpd, fang2020graspnet} by giving different friction coefficients $\mu$. By doing so, we will get a binary metric to indicate whether each grasp can successful grab the object under the friction coefficient $\mu$. $\mathrm{AP}_{\mu}$ means the average success rate of the predicted top-k grasps with k ranging from $1$ to $50$ under $\mu$, and $\mathrm{AP}$ is the average of $\mathrm{AP}_{\mu}$ for $\mu$ ranging from $0.2$ to $1.0$ with $\bigtriangleup \mu =0.2$.

\vspace{-5mm}
\subsubsection{Implementation Details.}
We now describe the implementation details of our EconomicGrasp framework. We use a 14-layers 3D UNet as our backbone, which is implemented by the Minkowski Engine\cite{choy20194d}. The output feature dimension is 512 and the hidden dimensions are [32, 64, 128, 256, 192, 192, 192, 192]. For the design of the loss function, we use the smooth L1 loss for view prediction. Regarding the angle, depths, widths and score, we use cross entropy for angle, depth and score learning and smooth L1 loss for width learning. In terms of the optimization, we use Adam\cite{kingma2014adam} as our optimizer with start learning rate $1e^{-3}$, and use the cosine learning rate decay schedule for training. The training batch size is 4 and the training epoch is 10. All the models are implemented with PyTorch\cite{paszke2019pytorch} on NVIDIA RTX3090 GPUs.

\begin{table}[t]
\centering
\caption{Performance comparison in Kinect camera. Best results are in bold face. We highlight the important comparison metrics by gray cell.}
\vspace{-2mm}
.\resizebox{1\textwidth}{!}{
\begin{tabular}{ccccccccccc}
\hline
\multirow{2}{*}{supervision} & \multirow{2}{*}{methods} & \multicolumn{3}{c}{seen} & \multicolumn{3}{c}{similar} & \multicolumn{3}{c}{novel} \\
\cmidrule(r){3-5} \cmidrule(lr){6-8} \cmidrule(l){9-11}
 & & \textbf{AP} & AP$_{0.8}$ & AP$_{0.4}$ & \textbf{AP} & AP$_{0.8}$ & AP$_{0.4}$ & \textbf{AP} & AP$_{0.8}$ & AP$_{0.4}$ \\
\hline
\multirow{2}{*}{sample} & GPD\cite{ten2017grasp} & \cellcolor{mygray}24.38 & 30.16 & 13.46 & \cellcolor{mygray}23.18 & 28.64 & 11.32 & \cellcolor{mygray}9.58 & 10.14 & 3.16 \\ 
& PointNetGPD\cite{liang2019pointnetgpd} & \cellcolor{mygray}27.59 & 34.21 & 17.83 & \cellcolor{mygray}24.38 & 30.84 & 12.83 & \cellcolor{mygray}10.66 & 11.24 & 3.21 \\ \hline
\multirow{3}{*}{sparse} & S4G\cite{qin2020s4g} & \cellcolor{mygray}18.72 & 22.90 & 14.58 & \cellcolor{mygray}11.82 & 14.74 & 7.67 & \cellcolor{mygray}5.38 & 6.69 & 2.19 \\ 
& TransGrasp\cite{liu2022transgrasp} & \cellcolor{mygray}35.97 & 41.69 & 31.86 & \cellcolor{mygray}29.71 & 35.67 & 24.19 & \cellcolor{mygray}11.41 & 14.42 & 5.87 \\  
& GraNet\cite{wang2023granet} & \cellcolor{mygray}41.38 & 49.84 & 33.86 & \cellcolor{mygray}35.29 & 43.15 & 26.89 & \cellcolor{mygray}11.57 & 14.31 & 5.24 \\ 
\hline
\multirow{3}{*}{dense} & GraspNet Baseline\cite{fang2020graspnet} & \cellcolor{mygray}29.88 & 36.19 & 19.31 & \cellcolor{mygray}27.84 & 33.19 & 16.62 & \cellcolor{mygray}11.51 & 12.92 & 3.56 \\
& TSB\cite{ma2023towards} & \cellcolor{mygray}49.42 & 58.38 & 43.66 & \cellcolor{mygray}41.49 & 50.29 & 34.60 & \cellcolor{mygray}15.35 & 19.18 & 7.98\\
& GSNet\cite{wang2021graspness} & \cellcolor{mygray}61.19 & 71.46 & \textbf{56.04} & \cellcolor{mygray}47.39 & 56.78 & 40.43 & \cellcolor{mygray}19.01 & 23.73 & 10.60 \\ \hline
economic & EconomicGrasp & \cellcolor{mygray}\textbf{62.59} & \textbf{73.89} & 55.99 & \cellcolor{mygray}\textbf{51.73} & \textbf{62.70} & \textbf{43.45} & \cellcolor{mygray}\textbf{19.54} & \textbf{24.24} & \textbf{11.12} \\
\hline
\end{tabular}}
\label{tab:kinect}
\vspace{-2mm}
\end{table}

\begin{table}[t]
\centering
\caption{Performance comparison in RealSense camera. Best results are in bold face. We highlight the important comparison metrics by gray cell.}
\vspace{-2mm}
\resizebox{1\textwidth}{!}{
\begin{tabular}{ccccccccccc}
\hline
\multirow{2}{*}{supervision} & \multirow{2}{*}{methods} & \multicolumn{3}{c}{seen} & \multicolumn{3}{c}{similar} & \multicolumn{3}{c}{novel} \\
\cmidrule(r){3-5} \cmidrule(lr){6-8} \cmidrule(l){9-11}
 & & \textbf{AP} & AP$_{0.8}$ & AP$_{0.4}$ & \textbf{AP} & AP$_{0.8}$ & AP$_{0.4}$ & \textbf{AP} & AP$_{0.8}$ & AP$_{0.4}$ \\
\hline
\multirow{2}{*}{sample} & GPD\cite{ten2017grasp} & \cellcolor{mygray}22.87 & 28.53 & 12.84 & \cellcolor{mygray}21.33 & 27.83 & 9.64 & \cellcolor{mygray}8.24 & 8.89 & 2.67 \\ 
& PointNetGPD\cite{liang2019pointnetgpd} & \cellcolor{mygray}25.96 & 33.01 & 15.37 & \cellcolor{mygray}22.68 & 29.15 & 10.76 & \cellcolor{mygray}9.23 & 9.89 & 2.74 \\ \hline
\multirow{3}{*}{sparse} & S4G\cite{qin2020s4g} & \cellcolor{mygray}25.71 & 31.19 & 20.84 & \cellcolor{mygray}18.45 & 23.14 & 12.86 & \cellcolor{mygray}9.04 & 11.23 & 3.61 \\ 
& TransGrasp\cite{liu2022transgrasp} & \cellcolor{mygray}39.81 & 47.54 & 36.42 & \cellcolor{mygray}29.32 & 34.80 & 25.19 & \cellcolor{mygray}13.83 & 17.11 & 7.67 \\  
& GraNet\cite{wang2023granet} & \cellcolor{mygray}43.33 & 52.56 & 34.04 & \cellcolor{mygray}39.98 & 48.66 & 32.00 & \cellcolor{mygray}14.90 & 18.66 & 7.76 \\ 
\hline
\multirow{3}{*}{dense} & GraspNet Baseline\cite{fang2020graspnet} & \cellcolor{mygray}27.56 & 33.43 & 16.95 & \cellcolor{mygray}26.11 & 34.19 & 14.23 & \cellcolor{mygray}10.55 & 11.25 & 3.98 \\
& TSB\cite{ma2023towards} & \cellcolor{mygray}58.95 & 68.18 & 54.88 & \cellcolor{mygray}52.97 & 63.24 & 46.99 & \cellcolor{mygray}22.63 & 28.53 & 12.00 \\
& GSNet\cite{wang2021graspness} & \cellcolor{mygray}65.70 & 76.25 & 61.08 & \cellcolor{mygray}53.75 & 65.04 & 45.97 & \cellcolor{mygray}23.98 & 29.93 & \textbf{14.05} \\ \hline
economic & EconomicGrasp & \cellcolor{mygray}\textbf{68.21} & \textbf{79.60} & \textbf{63.54} & \cellcolor{mygray}\textbf{61.19} & \textbf{73.60} & \textbf{53.77} & \cellcolor{mygray}\textbf{25.48} & \textbf{31.46} & 13.85 \\
\hline
\end{tabular}}
\label{tab:realsense}
\vspace{-4mm}
\end{table}

\begin{table}[t]
\centering
\caption{Training resource costs comparison. Best results are in bold face.}
\vspace{-2mm}
\resizebox{0.85\textwidth}{!}{
\begin{tabular}{ccccccc}
\hline
\multirow{2}{*}{supervision} & \multirow{2}{*}{methods} & \multicolumn{3}{c}{training} & mAP $\uparrow$  & mAP $\uparrow$  \\
\cmidrule(r){3-5}
& & time (h)$\downarrow$  & memory (G)$\downarrow$ & GPUs (G)$\downarrow$ & Kinect & RealSense \\
\hline
\multirow{2}{*}{sparse} & TransGrasp\cite{liu2022transgrasp} & 41.2 & 4.8 & 7.61 & 25.69 & 29.48  \\  
& GraNet\cite{wang2023granet} & 75.6 & 48.2 & 8.02 & 29.41 & 35.63  \\ 
\hline
\multirow{2}{*}{dense} & TSB\cite{ma2023towards} & 63.6 & 51.4 & 16.21 & 35.42 & 48.27  \\
& GSNet\cite{wang2021graspness} & 37.8 & 35.4 & 9.15 & 42.53 & 47.81 \\ 
\hline
economic & EconomicGrasp & \textbf{8.3} & \textbf{4.2} & \textbf{5.81} & \textbf{44.62} & \textbf{51.63} \\
\hline
\end{tabular}}
\label{tab:cost}
\vspace{-4mm}
\end{table}

\vspace{-3mm}
\section{Experiments}
\label{sec:exp}
\vspace{-3mm}

\subsection{Grasp Performance}
\vspace{-1mm}
\label{sec:performance}
To evaluate the performance of our economic framework, we conduct performance study comparing with other 6-DoF grasping methods, including GPD\cite{ten2017grasp}, PointNetGPD\cite{liang2019pointnetgpd}, S4G\cite{qin2020s4g}, GraspNet-Baseline\cite{fang2020graspnet}, TransGrasp\cite{liu2022transgrasp}, GraNet\cite{wang2023granet}, TSB\cite{ma2023towards}, GSNet\cite{wang2021graspness}. Among them, 
GPD\cite{ten2017grasp}, PointNetGPD\cite{liang2019pointnetgpd} are sample-based methods that trains by sampling different grasps and evaluating its score. S4G\cite{qin2020s4g}, TransGrasp\cite{liu2022transgrasp} and GraNet\cite{wang2023granet} are sparse supervision methods that trains on a partial of the GraspNet-1billion dataset, which output one or few grasps for each point. GraspNet-Baseline\cite{fang2020graspnet}, TSB\cite{ma2023towards} and GSNet\cite{wang2021graspness} are dense supervision methods that trains on dense labels, which learns the grasp from all the $(300$ view $\times12$ rotation $\times4$ depth) grasp labels for each points.

The testing results of Kinect and RealSense cameras are shown in Fig. \ref{tab:kinect} and Fig. \ref{tab:realsense} respectively. From the table we can see that our EconomicGrasp outperforms all the SOTA methods. For example, we outperform GSNet\cite{wang2021graspness} by 1.40, 4.32, 0.53 AP increases on seen, similar and novel scenes in Kinect camera, and by 2.51, 7.42, 1.50 AP increases on seen, similar and novel scenes in RealSense camera. On average, there is about 3 AP enhancement, showing great potential for the economic framework. The improvement could be attributed to the following reasons: 1) our economic framework innovatively maintains grasp views for training, which mitigates the ambiguity problem; 2) our focal representation module is effective and specifically designed for our economic supervision. The result shows great potential for the economic framework.

\vspace{-3mm}
\subsection{Training Resource Cost}
\vspace{-2mm}
\label{sec:training_cost}
To evaluate the economy of our economic framework, we also conduct the training costs experiments comparing with some effective 6-DoF grasping methods, including TransGrasp\cite{liu2022transgrasp}, GraNet\cite{wang2023granet}, TSB\cite{ma2023towards} and the current SOTA dense supervision method GSNet\cite{wang2021graspness}. Specifically, we test the resources of total training time, memory costs and GPU costs for training the GraspNet-1Billion dataset on Kinect data. All the experiments are conducted in an empty machine without other running processes for fair comparison.

\begin{table}[t]
\begin{minipage}{0.7\textwidth}
\centering
\caption{Real world experiments. Best is in bold face.}
\resizebox{0.9\textwidth}{!}{
\begin{tabular}{cccccc}
\toprule
\multirow{2}{*}{IDs} & \multirow{2}{*}{objects} & \multicolumn{2}{c}{GSNet\cite{wang2021graspness}} & \multicolumn{2}{c}{EconomicGrasp} \\
  &  & attempts & success & attempts & success \\
\midrule
1, 4, 9, 11, 15 & 5 & 6 & 83.3\% & 5 & 100\% \\
2, 3, 6, 7, 12 & 5 & 5 & 100\% & 6 &  83.3\% \\
5, 8, 10, 11, 13, 14 & 6 & 7 & 85.2\% & 7 & 85.2\% \\
1, 2, 3, 11, 12, 15 & 6 & 7 & 85.2\% & 6 & 100\% \\
5, 6, 7, 9, 13, 14, 15 & 7 & 8 & 87.5\% & 8 & 87.5\% \\
1, 3, 6, 8, 10, 11, 12 & 7 & 8 & 87.5\% & 7 & 100\% \\
total & 36 & 41 & 87.8\% & \textbf{39} & \textbf{92.3\%} \\
\bottomrule
\end{tabular}}
\label{tab:real_world}
\end{minipage}
\begin{minipage}{0.29\textwidth}
\centering
\caption{Analysis of bad cases. It shows the number of testing scenes without any successful grasps in the generated top50 grasps. There are 7680 scenes in total. Best results are in bold face.}
\resizebox{0.95\textwidth}{!}{
\begin{tabular}{cc}
\toprule
 GSNet\cite{wang2021graspness} & EconomicGrasp \\
\midrule
109/7680 & \textbf{36}/7680 \\
\bottomrule
\end{tabular}}
\label{tab:fail_cases}
\end{minipage}
\vspace{-2mm}
\end{table}

As shown in Table \ref{tab:cost}, we can see that our method has the lowest training resource costs in training time, memory utilization and GPU utilization comparing to other methods, meanwhile surpasses all methods in grasp performance. For example, comparing to the SOTA method GSNet\cite{wang2021graspness}, our method has 3 AP improvement on average in performance, and only uses 1/4 training time, 1/8 memory costs. This is due to our economic framework that using less but important supervision to train the model, and design a tailored focal representation module for the economic training.

\begin{table}[t]
\begin{minipage}{0.6\textwidth}
\centering
\caption{Main ablation study. Important comparison metric is in gray cell.}
\resizebox{0.95\textwidth}{!}{
\begin{tabular}{ccccccc}
\hline
economic & interactive & composite & \multirow{2}{*}{seen} & \multirow{2}{*}{similar} & \multirow{2}{*}{novel} & \multirow{2}{*}{\textbf{mean}} \\ 
supervision & head & score & & & \\ 
\hline
 &  & & 43.59 & 34.09 & 13.36 & \cellcolor{mygray}30.34 \\
 \checkmark &  &  & 60.07 & 48.16 & 18.70 &  \cellcolor{mygray}42.31 \\
 \checkmark & \checkmark  &  & 63.08 & 50.61 & 18.74 & \cellcolor{mygray}44.14  \\
 \checkmark &   & \checkmark & 59.81 & 48.45 & 19.01 & \cellcolor{mygray}42.42  \\
 \checkmark & \checkmark  & \checkmark & 62.59 & 51.73 & 19.54 & \cellcolor{mygray}44.62  \\
\hline
\end{tabular}}
\label{tab:main_ablation}
\end{minipage}
\begin{minipage}{0.4\textwidth}
\centering
\caption{Ablation study of the economic supervision paradigm.  Important comparison metric is in gray cell.}
\resizebox{0.95\textwidth}{!}{
\begin{tabular}{cccccc}
\hline
selective & \multirow{2}{*}{seen} & \multirow{2}{*}{similar} & \multirow{2}{*}{novel} & \multirow{2}{*}{\textbf{mean}} \\ 
loss & & & & \\ 
\hline
& 59.67 & 48.68 & 19.27 & \cellcolor{mygray}42.54 \\
\checkmark  & 62.59 & 51.73 & 19.54& \cellcolor{mygray}44.62 \\
\hline
\end{tabular}}
\label{tab:abla_eco_supervision}
\end{minipage}
\end{table}

\begin{table}[t]
\begin{minipage}{0.5\textwidth}
\centering
\caption{Ablation study of the interactive grasp head. Important comparison metric is in gray cell.}
\resizebox{0.95\textwidth}{!}{
\begin{tabular}{cccccc}
\hline
global  & local & \multirow{2}{*}{seen} & \multirow{2}{*}{similar} & \multirow{2}{*}{novel} & \multirow{2}{*}{\textbf{mean}} \\ 
interactive & interactive & & & &\\ 
\hline
 &  & 59.81 & 48.45 & 19.01 & \cellcolor{mygray}42.42\\
\checkmark &  & 61.96 & 49.09 & 19.44 & \cellcolor{mygray}43.49 \\
\checkmark & \checkmark  & 62.59 & 51.73 & 19.54 & \cellcolor{mygray}44.62\\
\hline
\end{tabular}}
\label{tab:abla_interactive}
\end{minipage}
\begin{minipage}{0.5\textwidth}
\centering
\caption{Ablation study of the composite score estimation. Important comparison metric is in gray cell.}
\resizebox{0.95\textwidth}{!}{
\begin{tabular}{cccccc}
\hline
\multirow{2}{*}{classification}  & composite & \multirow{2}{*}{seen} & \multirow{2}{*}{similar} & \multirow{2}{*}{novel} & \multirow{2}{*}{\textbf{mean}} \\ 
 & estimation & & & &\\ 
\hline
 &  & 63.08 & 50.61 & 18.74 & \cellcolor{mygray}44.14 \\
\checkmark &  & 51.66 & 38.71 & 14.58 & \cellcolor{mygray}34.98 \\
\checkmark & \checkmark  & 62.59 & 51.73 & 19.54 & \cellcolor{mygray}44.62 \\
\hline
\end{tabular}}
\label{tab:abla_composite_score}
\end{minipage}
\vspace{-2mm}
\end{table}

\begin{figure}[t]
  \centering
  \includegraphics[width=1\textwidth]{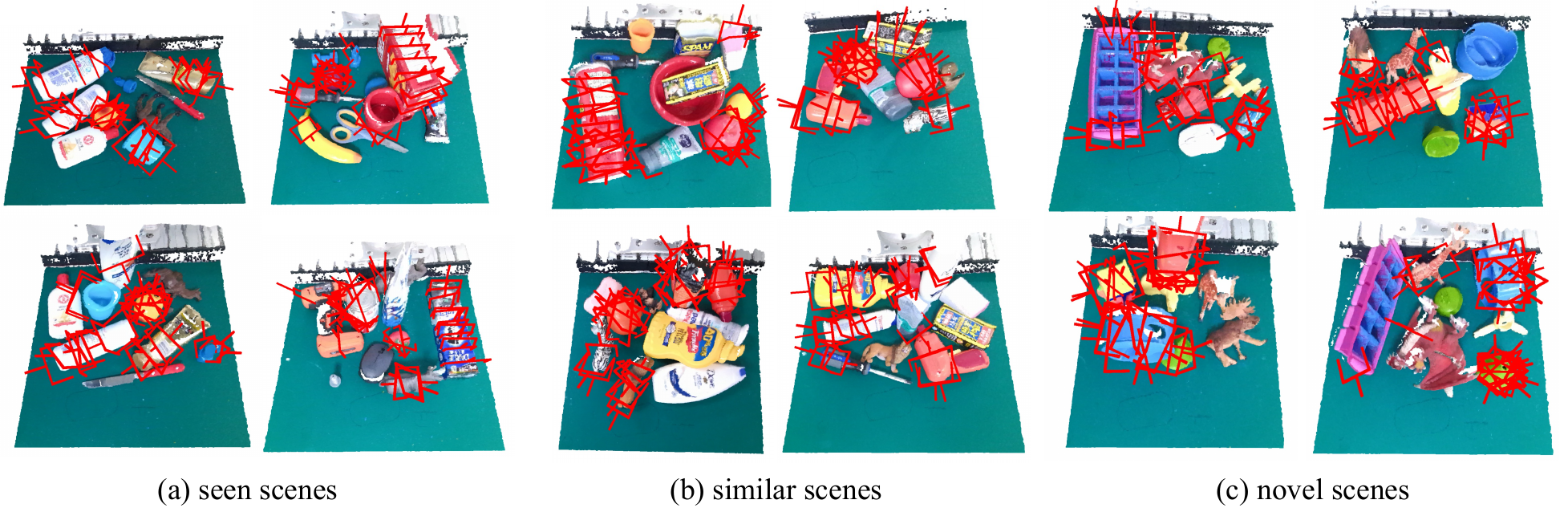}
  \vspace{-5mm}
  \caption{Qualitative analysis.}
  \vspace{-6mm}
  \label{fig:qualitative}
\end{figure}

\vspace{-3mm}
\subsection{Ablation Study}
\label{sec:ablation}
\vspace{-2mm}
Moreover, we conduct a series of ablation studies to test the effectiveness of each design of our framework. All the experiments are trained on GraspNet-1Billion dataset\cite{fang2020graspnet} on the Kinect data.

\vspace{-3mm}
\subsubsection{Main Ablation Study.} We first conduct ablation study to test the main components of our framework. As shown in the first line in Table \ref{tab:real_world}, the economic supervision paradigm gains the major improvements for our method, since it mitigates the key ambiguity problem when reducing the supervision. Moreover, from the second line in  Table \ref{tab:real_world}, we can see that interactive head also helps to enhance the performance by 1.8 mean AP, showing that interactively learning focal and discriminate features for a specific grasp is useful. However, as shown in the third line in Table \ref{tab:real_world}, the composite score estimation cannot amplify the performance along. 
But with the help of the interactive head to learn focal representation for the specific grasp, the composite score can be used together with the interactive grasp head and improve the performance, as shown in the forth line in Table \ref{tab:real_world}. Due to the space limit, the ablation study for resource costs of the main components is put in the supplementary material.

\vspace{-3mm}
\subsubsection{Other Ablation Studies.} We also conduct other ablation studies for each component. As shown in Table \ref{tab:abla_eco_supervision}, the selective loss can improve the model performance with about 2.08 AP, indicating that masking the unmatched points for loss calculation is important for training under economic supervision. In addition, the ablation study for the interactive grasp head is shown in Table \ref{tab:abla_interactive}. It demonstrates that by gradually adding the global interactive attention and the local interactive attention, the model performance can be elevated step by step, showing that learning interaction to focus on a specific grasp is effective for economic supervision. Furthermore, the ablation study of the composite score estimation is demonstrated in Table \ref{tab:abla_composite_score}. From the table, we can observe that composite score estimation at inference is necessary for predicting accurate scores. Predicting the score without it will cause nearly 10 AP decrease. We also conduct several ablation studies about keeping different numbers of views and comparing different types of sparse supervision, which can be seen in the supplementary material.

\vspace{-3mm}
\subsection{Real-world Testing}
\label{sec:real_world}

\begin{wrapfigure}{r}{0.6\textwidth}
  \centering
  \vspace{-15mm}
  \includegraphics[width=0.6\textwidth]{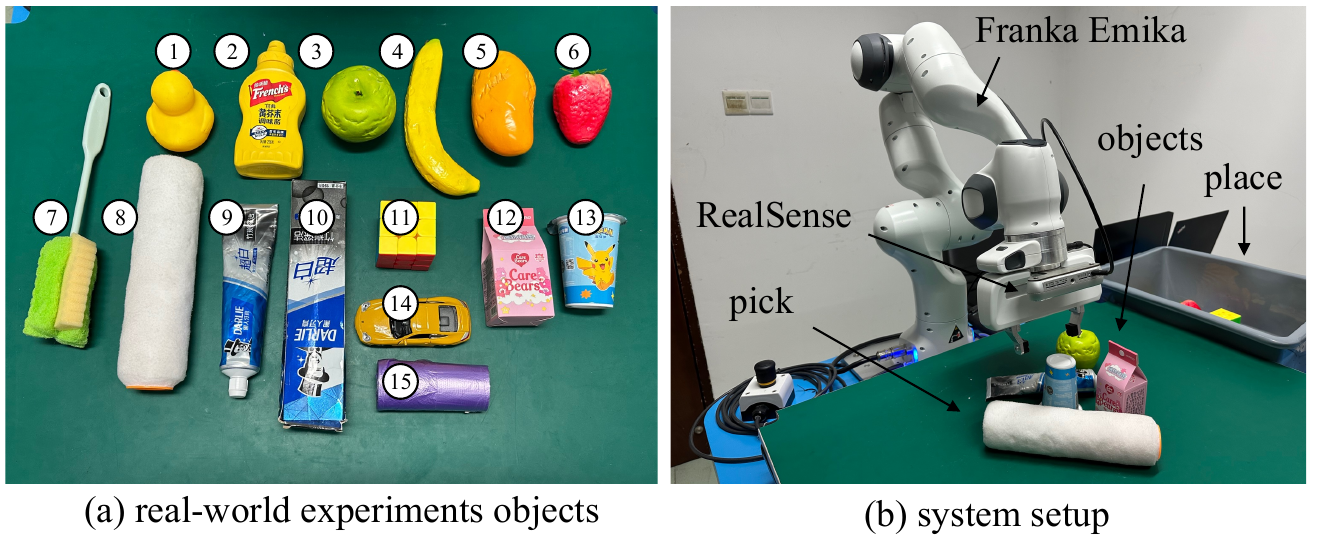}
  \caption{Real-world experiments setup.}
  \label{fig:real_world}
  \vspace{-8mm}
\end{wrapfigure}

For a deeper assessment of our performance, we further conduct the real-world experiments comparing with the SOTA method GSNet\cite{wang2021graspness}. In detail, we arrange six scenes with 5,5,6,6,7,7 objects and test the number of grasps required to empty the scenes (a pick and place task). The experiments are conducted by Franka Emika robot hand with a two-finger gripper. The experiment setup and the testing objects are shown in Fig. \ref{fig:real_world}, which contains 4 trained objects (NO. 3, 4, 6, 10) and 11 untrained objects (NO. 1, 2, 5, 7, 8, 9, 11, 12, 13, 14, 15). The testing results are shown in Table \ref{tab:real_world}. It can be seen that our method can achieve robust real-world grasping ability, slightly surpassing the SOTA method GSNet\cite{wang2021graspness} in the real world experiments, showing great potential for our economic framework. We also make a video demo, showing the real world grasping practice with our economic framework, which can be found in the supplementary material.

\vspace{-3mm}
\subsection{Analysis}
\subsubsection{Qualitative Analysis.}
We conduct qualitative analysis to further evaluate the robustness and effectiveness of our method. We randomly sample some scenes from the test set of GraspNet-1Billion dataset\cite{asif2018graspnet} and visualize the top30 grasps generated by our method. The results are shown in Fig. \ref{fig:qualitative}. From the figure, we can see that the grasps generated by our method are robust and basically reasonable, demonstrating the effectiveness of our economic framework. In addition, the generated grasp poses are diverse and flexible, showing the effectiveness of 6-DoF grasping. We also test the diversity of our generated grasp in the supplementary material to show that we maintain the diversity although we reduce the dataset.

\vspace{-3mm}
\subsubsection{Failure Cases Analysis.}
Additionally, we also calculate the number of failure cases of our model comparing with the SOTA method GSNet\cite{wang2021graspness}. To be specific, we analyze the number of scenes in the seen test set that has no successful grasps in the predicted top50 grasps with low  friction coefficients $\mu=0.2$, which can evaluate the high-quality grasps generating ability of our method.

As demonstrated in Table \ref{tab:fail_cases}, we can see that in the total of 7680 scenes, there are 109 failure cases for GSNet\cite{wang2021graspness}, and only 36 failure cases for our model, which indicates that selecting high-quality labels to supervise the learning is to some extent robust to generate high-quality grasps, which is also showing the great potential for the exploration of economic grasping. We also conduct plug-and-play experiments in the supplementary material, showing great potential for our economic procedure.

\vspace{-3mm}
\section{Conclusion}
\label{sec:conclusion}
\vspace{-2mm}
In this work, we propose an economic framework for 6-DoF grasp detection, to economize resource costs in training and maintain effective grasp performance. We find that the dense supervision is the bottleneck to encumber the entire training overload, and meanwhile make the training hard to converge. Toward this end, we propose the economic supervision paradigm to reduce the supervision and meanwhile maintain comparable performance. Furthermore, economic supervision gives us the chance to focus on the learning of the specific grasp, thus we devise a focal representation module to enable it. Combining all together, we propose the EconomicGrasp framework. Our extensive experiments show that our framework surpasses the SOTA grasping method with low resources cost (1/4 training time cost, 1/8 memory cost and 1/25 storage cost), showing great potential for the economic framework. 

In the future, exploring how to build economic grasping supervision from natural, instead of simplifying from a dense dataset, will be interesting and meaningful, which can be the future work to do. And we believe this work can also give insights to the further exploration about it.

\clearpage  

\section*{Acknowledgements}
This work was supported partially by the National Key Research and Development Program of China (2023YFA1008503), NSFC(U21A20471), Guangdong NSF Project (No. 2023B1515040025, 2020B1515120085). Additionally, I sincerely thank Jia-Run Du's help for the valuable suggestions for the paper.

%
%
\bibliographystyle{splncs04}
\bibliography{main}
\end{document}